\documentclass[10pt,journal,compsoc]{IEEEtran}

\usepackage{refstyle}
\usepackage{titlesec}
\usepackage{hyperref}
\usepackage{cleveref}
\usepackage{subfig}
\usepackage{eurosym}
\usepackage{xcolor}
\usepackage{graphicx}
\usepackage[per-mode=symbol]{siunitx}            
\usepackage{mathptmx}   
\usepackage{geometry}   
\usepackage[backend=biber,      
    style=ieee,
    bibstyle=ieee,              
    citestyle=numeric-comp,     
    maxbibnames = 3,
    sortcites=true,   
    giveninits=true,           
    backref=false,
    doi=true,                  
    url=false,                  
    isbn=false,                 
    uniquename = false,
    uniquelist = false
]{biblatex}
\newcommand{\BCref}[1]{\textbf{\Cref{{#1}}}}    
\titleformat{\section}
  {\normalfont\fontsize{12}{17}\bfseries}
  {\thesection}
  {1em}
  {}

\titleformat{\subsection}
  {\normalfont\fontsize{12}{17}\bfseries}
  {\thesubsection}
  {1em}
  {}

\begin{document}

\captionsetup[figure]{labelfont={bf},name={Figure},labelsep=period}

\title{\raggedright \bf \Large Upside down: affordable high-performance motion platform}

\author{\raggedright \small 
        Nayan~Man~Singh~Pradhan$^1$,
        Patrick Frank$^1$,
        An Mo$^{1,*}$, and
        Alexander Badri-Spr\"owitz$^{1,2}$\\
        {$^1$Dynamic Locomotion Group, Max Planck Institute for Intelligent Systems, Stuttgart, Germany}\\
        {$^2$Department of Mechanical Engineering, KU Leuven, Belgium}\\
        *\texttt{mo@is.mpg.de}
}

\IEEEtitleabstractindextext{%
\begin{abstract}

Parallel robots are capable of high-speed manipulation and have become essential tools in the industry.
The proximal placement of their motors and the low weight of their end effectors make them ideal for generating highly dynamic motion. 
Therefore, parallel robots can be adopted for motion platform designs, as long as end effector loads are low.
Traditional motion platforms can be large and powerful to generate multiple g acceleration. However, these designs tend to be expensive and large. 
Similar but smaller motion platforms feature a small work range with reduced degrees of freedom (DoFs) and a limited payload.
Here we seek a medium-sized affordable parallel robot capable of powerful and high-speed 6-DoF motion in a comparably large workspace.
This work explores the concept of a quadruped robot flipped upside-down, with the motion platform fixed between its feet. 
In particular, we exploit the high-power dynamic brushless actuation and the four-leg redundancy when moving the motion platform.
We characterize the resulting motion platform by tracking sinusoidal and circular trajectories with varying loads.
Dynamic motions in 6 DoFs up to 10 Hz and $\pm$ 10 mm amplitude are possible when moving a mass of 300 grams. 
We demonstrate single-axis end-effector translations up to $\pm$ 20 mm at 10 Hz for higher loads of 1.2 kg. 
The motion platform can be replicated easily by 3D printing and off-the-shelf components.
All motion platform-related hardware and the custom-written software required to replicate are open-source. 


\end{abstract}

}

\maketitle
\IEEEdisplaynontitleabstractindextext
\IEEEpeerreviewmaketitle

\section{Introduction}
\label{sec:introduction}

Motion platforms are designed to move objects in 3D space in a given workspace. 6-degree of freedom (DoF) motion platforms are widely adapted to simulate the complex and dynamic motions of aircraft, vehicles, ships, entertainment platforms, and research test benches~\cite{wei2021design,li2019design}.

Multiple approaches for generating 6-DoF motion have been proposed, such as using industrial 6 DoF robotic arms~\cite{iqbal2012modeling}, crank arm platforms~\cite{wei2021design,fomin2021inverse}, linear actuators~\cite{faulring2004high,fujimoto1990development}, and parallel robots.
Parallel robots are \textit{"closed-loop mechanical structures whose mobile platforms are linked to the base by independent kinematic chains, presenting good potential in terms of accuracy, rigidity, and ability to manipulate large loads with positioning errors"}~\cite{staicu2019dynamics}.
Thanks to their high rigidity, robustness against external force and good payload abilities\cite{gao2005generalized,ghobakhloo2006position,zhang2019forward,ratiu2020brief}, parallel robots are often adopted for high acceleration motion platform designs.

Delta robots~\cite{pierrot1990delta,rey1999delta,bonev2001delta} and Stewart-Gough platforms~\cite{gao2005generalized,ghobakhloo2006position} are well known solutions. Delta robots feature three translational DoFs where the end-effector stays parallel to the base of the robot, allowing for high-speed pick and place tasks~\cite{bonev2001delta,merlet2005parallel,mcclintock2018millidelta}. Stewart-Gough platforms consist of a platform connected to a fixed base using six extensible legs of adjustable lengths, which determine the position and orientation of the attached platform~\cite{gao2005generalized}.
Despite their popularity, such robots are still costly and less well suited for non-industrial applications.


High performance 6-DoF motion platforms are more expensive with price tags above 100k \euro, as they are intended for simulating the motion of large objects. These commercial motion platforms are operated  with proprietary, custom-designed software.
We found that limited affordable options for motion simulation exist for research.
Low-end motion platforms are often designed for education purposes, and deliver insufficient acceleration in case of high-range motions and higher mass objects.
Hence, researchers have developed motion platforms from off-the-shell components~\cite{mo2021open}.
Clearly, building an entire motion platform and its software framework from scratch is time-consuming.

\begin{figure}[tbp]%
    \centering
    \subfloat[\centering Simulation environment. \label{fig:pybullet_solo_env}]{{\includegraphics[width=0.4\linewidth]{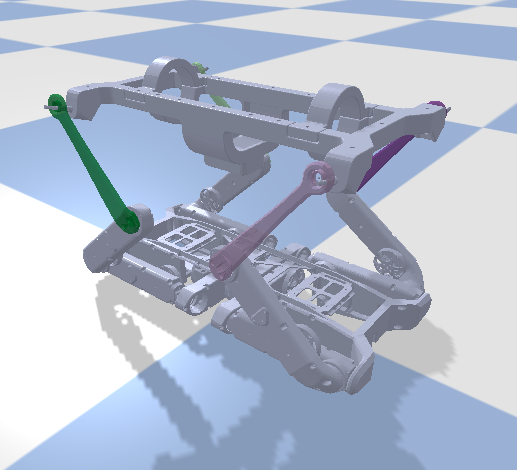} }}%
    \qquad
    \subfloat[\centering Hardware photo. \label{fig:solo_solo_env}]{{\includegraphics[width=0.4\linewidth]{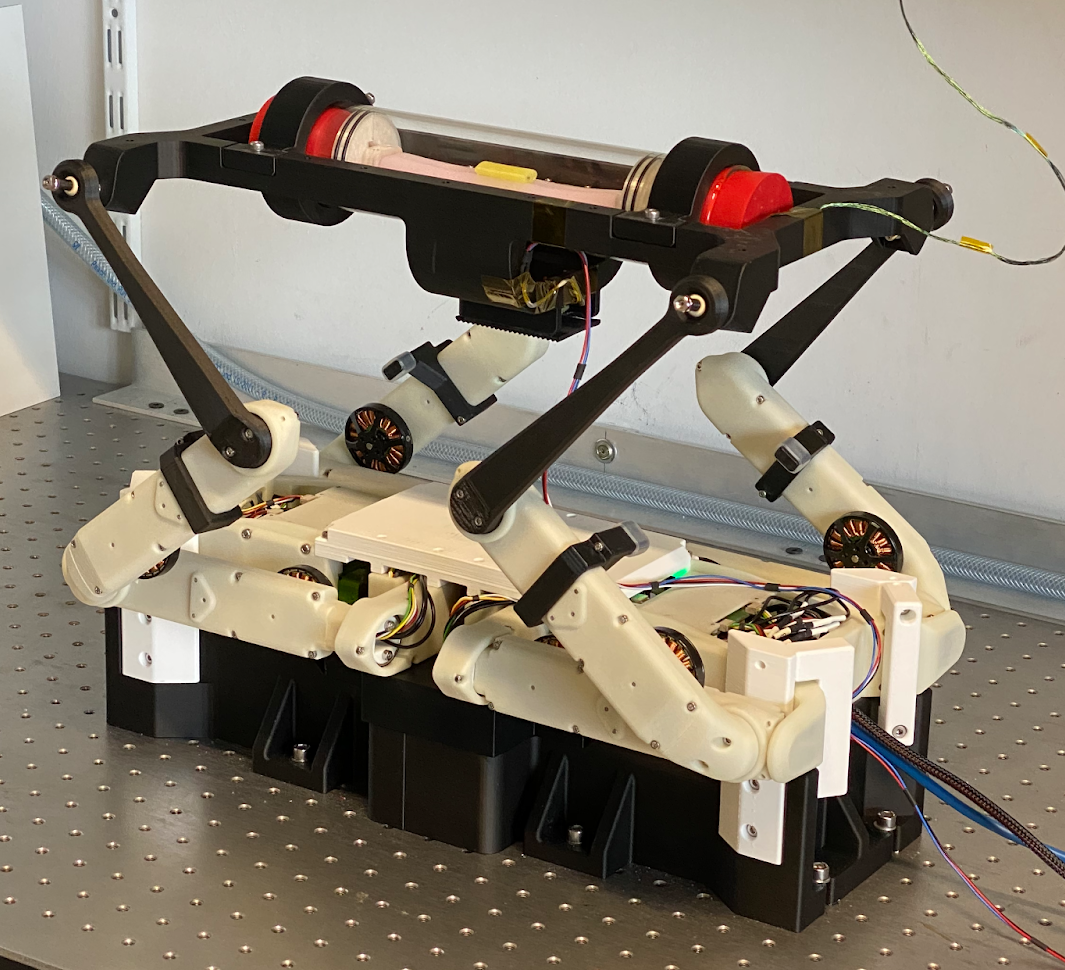} }}%
    \qquad
    \caption{The proposed 6-DoF motion platform. The right figure shows a high-weight load (1.2 kg) mounted to the motion platform.} %
    \label{fig:control_envs}%
\end{figure}

Here, we were looking for open-source robot platforms as solutions to build an affordable, customized motion platform (\BCref{fig:control_envs}).
For our specific need of high acceleration motion simulation, we could not find designs at affordable prices.
But as a robotic research laboratory, we work daily with dynamically running legged robots that are open-source available.
Consequently, we build our solution around the recently open-sourced quadruped robot Solo~\cite{grimminger2020open}.
Solo robot is capable of highly dynamic movements, such as high vertical jumps, and even back flips.
Its highly dynamic robot motion is made possible by the robot's powerful brushless motors controlled by field oriented control (FOC).
By flipping the robot upside down, its four legs can be used to actuate a motion platform in 6-DoF.
The four-legged SOLO robot (SOLO-12) features in sum twelve motors with 3 motors per leg; two shoulder actuators, and one knee actuator. 
A comparably low gearbox ratio (9:1) allows for high accelerations, and powerful brushless motors supply high torque output.
The Solo robot is designed with off-the-shell components and made mostly from 3D printing parts, which lead to comparably low production and maintenance costs.


In this work, we present the resulting high-performance and affordable 6-DoF motion platform hardware design and its software and control framework. 
Our hardware design is based on the quadruped robot Solo-12. 
As we flip the Solo robot upside down, making the legs face upwards and attach to our custom designed platform, we created a parallel robot with closed-loop kinematic chains.
We developed a multibody simulation model of the robot in the PyBullet environment~\cite{coumans2021}, which we connect with our robot control framework. %
Both hardware design and software framework are open-source available. 
The proposed motion platform achieves highly dynamic movement.
It tracks a sinusoidal trajectory up to $10$ Hz with  $\pm 10$ mm translation amplitude, for motion platform loads of 0.3 kg. %
Higher loads of 1.2 kg can be moved at 2 Hz, for a motion range of $\pm 20$ mm translation or $\pm 10$ degrees rotation. %
To the best of our knowledge, there is no open-source 6-DoF motion platform capable of such highly dynamic movement, with a comparable price tag.


\section{Robot Framework}

\subsection{Hardware Framework}

\begin{figure}[htp]
    \centerline{\includegraphics{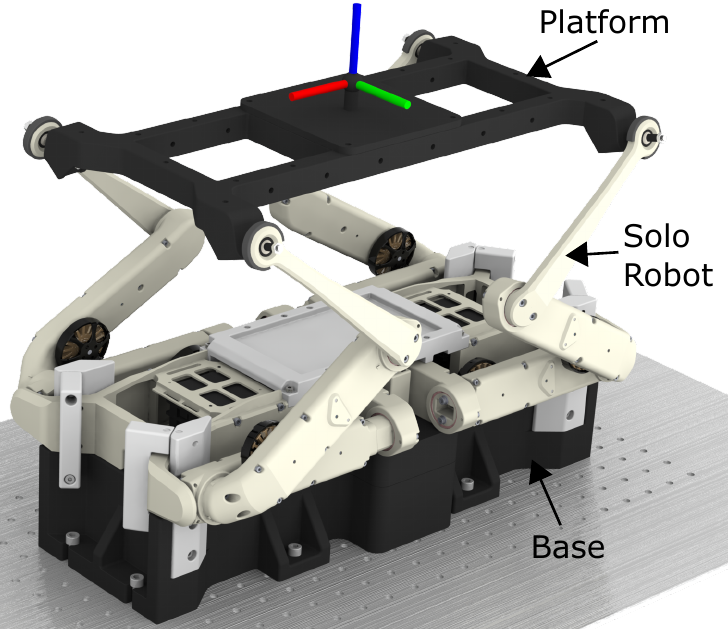}}
    \caption{Main components of the motion platform. The coordinate system on the platform shows x, y, z directions in red, green and blue.}
    \label{fig:solo_robot_labelled}
\end{figure}

The hardware framework consists of three main components (\BCref{fig:solo_robot_labelled}); a quadruped robot Solo~\cite{grimminger2020open}, a light-weight motion platform attached to the robot's legs, and a base and fixture to mount the SOLO robot body to the table. 
The complete setup requires a total installation space with a height of $50$ cm (z-axis), length of $110$ cm (x-axis), and a width of $80$ cm (y-axis). Safety margins are included.
The maximum workspace of the platform is x-axis $\pm$ \SI{255}{mm},  y-axis $\pm$ \SI{105}{mm},  and z-axis $\pm$ \SI{105}{mm} for translation; 30 degrees for roll, yaw and pitch.

The Solo robot~\cite{grimminger2020open} is an open source modular robot. The SOLO actuator module features a $9:1$ dual-stage timing belt transmission, a high-resolution optical encoder, and a brushless motor. Each SOLO joint can output $\tau_{max}=2.7$ Nm joint torque, and is controlled at up to \SI{1}{kHz}~\cite{grimminger2020open}. 

The platform is mounted to the robot's feet with ball joints (IGUS KGLM-05). 
Ball joint allow a maximum pivot angle of 30 degrees. 
We press-fit the ball joints into the robot's feet and secured them with M4 nuts to the platform. 
The platform itself is 3D printed from PLA material. 
The platform provides multiple mounting points for attaching payloads. 
An inertia measurement unit (IMU, 3DM-CX5-25) is mounted at the bottom of the platform.
With the IMU, we record end effector ground truth data.
To fix the robot body during movement, we designed a base that is screwed on top of an optic table. 
We replaced the lower center plate of the original robot with our mounting plate. 
CAD models of the platform and the base are available on GitHub \url{https://github.com/nayan-pradhan/solo-6dof-motion-platform}.
The cost for the 3D printing parts of the motion platform's base, clamps, the platform, and the connecting ball joints amounts to approximately $100$ \euro. %
The material cost of the SOLO robot is approximately $6,300$ \euro \ without an inertial measurement unit (IMU), and  $8,000$ \euro \ with an IMU (Lord Microstrain 3DM-CX5-25). 

\subsection{Software Framework}
We custom-wrote a software framework to control platform motion. This open-source software package offers full control for 6 DoF motion. In addition to the physical robot control, we also offer a PyBullet-based simulation environment to evaluate the performance of the platform. The framework features two options of logging the platform pose - through the physically mounted IMU or through forward kinematic calculations of joint angles. Because of the high price of higher quality IMUs, the forward kinematic calculation is the low-cost option.
The software package is mainly written in Python3. The software framework is built modular such that users may update, add, or remove parts of the module as long as the input format is maintained. The detailed documentation can be found in our GitHub repository: \url{https://github.com/nayan-pradhan/solo-6dof-motion-platform}.

\begin{figure}[htbp]
    \centerline{\includegraphics{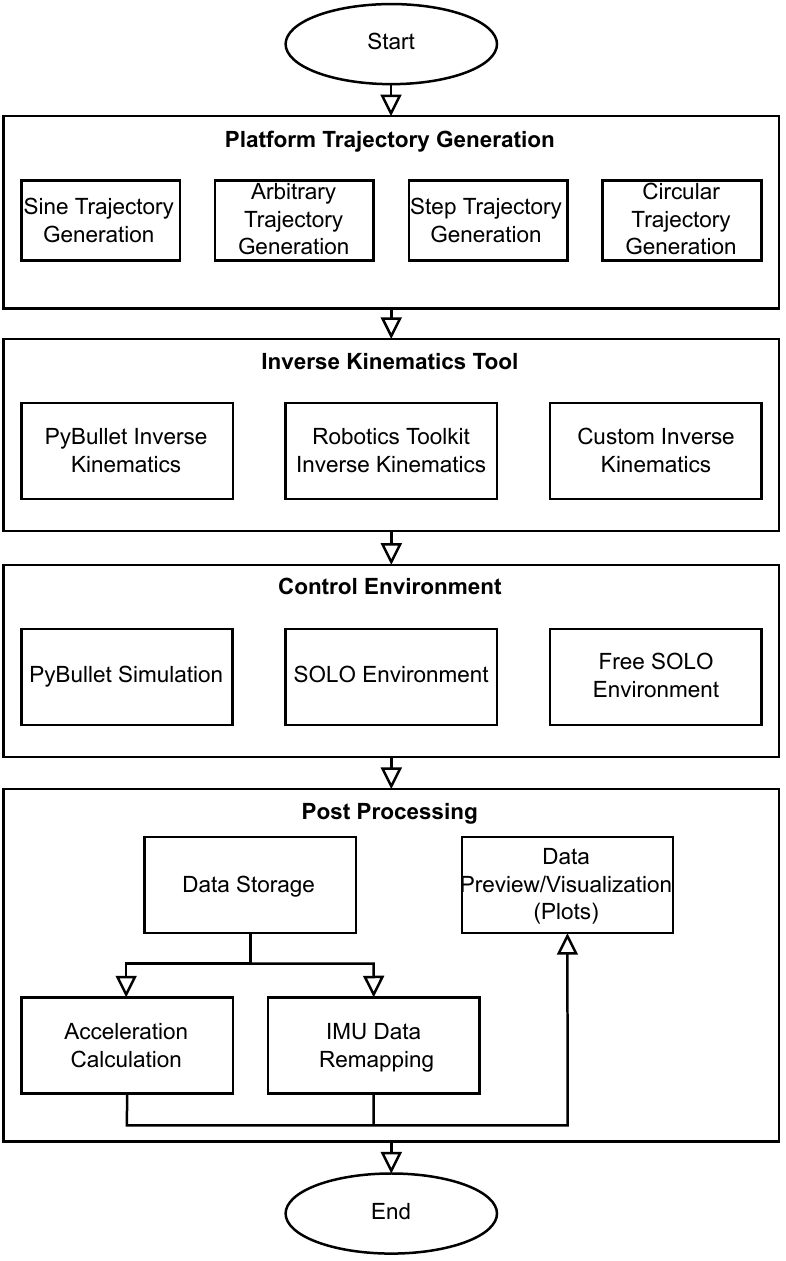}}
    \caption{Software framework.}
    \label{fig:flowchart_software}
\end{figure}

Our software framework consists of four main modules, shown in \BCref{fig:flowchart_software}:
\begin{enumerate}
    \item Platform Trajectory Generation
    \item Inverse Kinematics Tool 
    \item Control Environment
    \item Post Processing
\end{enumerate}

\subsubsection{Platform Trajectory Generation}
\label{sec:platform_trajectory_generation}
The Platform Trajectory Generation module generates the target pose of the platform. We implemented four base trajectory options:

\textbf{Sine Trajectory}: A sine wave trajectory of specified run time (seconds), wait time (seconds), type of movement (translation/rotation), frequency (Hz), amplitude (mm/deg), and x, y, z axis offsets (mm) is generated.

\textbf{Arbitrary Trajectory}: A smooth linear interpolation of specified time (seconds) between a series of arbitrary target positions (mm) and orientations (deg) is generated. 

\textbf{Step Trajectory}: A step function in a specified position (mm) and orientation (deg) is generated. As such, a step response is useful for tuning controller gains. %

\textbf{Circular Trajectory}: A circular trajectory of specified radius of translation (mm), angle of rotation (deg), number of rounds, frequency of rotation, direction of translation and rotation (clock-wise/counter-clock-wise), and ability to enable/disable translation and/or rotation is generated. 


\subsubsection{Inverse Kinematics Tool}
\label{sec:inverse_kinematics_tool}
The Inverse Kinematics Tool module takes platform target trajectories generated from \BCref{sec:platform_trajectory_generation} as the input and calculates target joint angles for each joint using inverse kinematics. We use PyBullet~\cite{pybullet_home_page} for the inverse kinematics calculation. Additionally, we provide an API for custom inverse kinematics packages such as Robotics Toolbox~\cite{rtb}. %

\subsubsection{Control Environment}
\label{sec:control_environment_section}
The Control Environment module takes the target joint angles generated from \BCref{sec:inverse_kinematics_tool} as the input and starts the platform motion. Our software framework contains two control environments:
\begin{enumerate}
    \item PyBullet Simulation Environment
    \item Solo Environment
\end{enumerate}
The PyBullet Simulation Environment (\BCref{fig:pybullet_solo_env}) executes joint commands in simulation at \SI{240}{Hz}. The control update frequency is limited by PyBullet. The simulation environment loads the URDF models of the Solo robot and the platform. It creates constraints between foot joints and platform corners. The PyBullet simulation can be used to test new algorithms and implementations without wearing the physical robot setup out.

The Solo Environment (\BCref{fig:solo_solo_env}) is used to control our physical motion platform at \SI{1000}{Hz}. The Solo environment initializes the robot, calibrates the robot, executes the desired motion trajectory, and logs the all sensor data.


\subsubsection{Post Processing}
\label{sec:post_processing}
The Post Processing module loads the stored data, processes data, and plots previews. 

Target trajectories and sensor readings from Sections \ref{sec:platform_trajectory_generation}, \ref{sec:inverse_kinematics_tool}, and \ref{sec:control_environment_section} are loaded. We calculate the pose of the platform from robot joint angles. %

We do this by first calculating the position of the foot links in the Solo robot (our end-effectrs) for each leg using the loaded joint angles and the measured distances between joints from the CAD model. We define a dummy ball joint at the foot links (end-effectors). We then construct a vector from the front left ball joint to the back right ball joint and another vector from the front right ball joint to the back left ball joint. We compute the intersection point between the two constructed vectors. This intersection point represents the x and y values of the platform center. In order to get the correct z value, we have to add an offset in z, which we get from the CAD model.

To get the orientation of the platform we calculate three more vectors. First we calculate the vector from the front left to the back left ball joint position. This vector represents the x direction of our platform. By calculating the vector from the front left to the front right we get the y direction. Lastly we get the z direction by building the cross product of the two vectors we calculated earlier. To get the platform orientation we use the Rotation class from \texttt{scipy.spation.transform}. We use the method \texttt{align\_vectors()} where we get the transformation matrix between the global coordinate system and our three calculated vectors representing the platform axes. This transformation matrix represents our platform orientation. For easier understanding we chose to use the Euler angle representation.

The linear and angular velocity and acceleration at the center of our platform is derived using the calculated positions and orientations. This provides the use the option to record the position, velocity, and acceleration of the platform without purchasing an expensive IMU. The calculated pose is filtered using a Butterworth low pass filter with a cutoff frequency of \SI{50}{Hz}. 

\begin{figure}[htbp]
    \centerline{\includegraphics{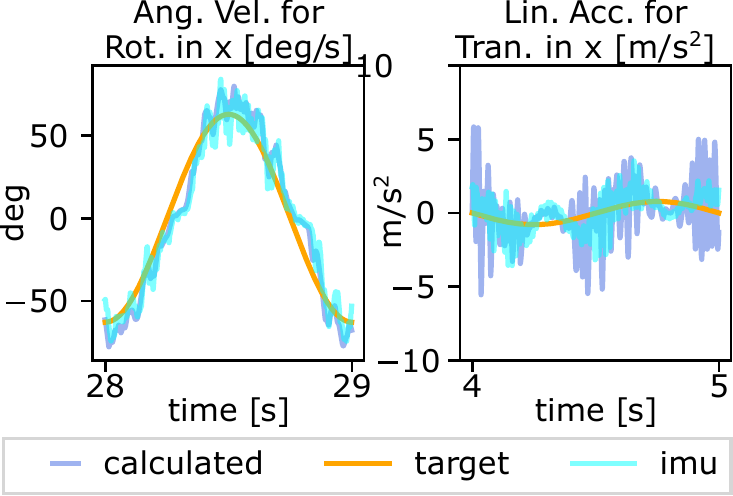}}
    \caption{Calculated, target, and IMU angular velocity and linear acceleration in x-direction.}
    \label{fig:imu_overlay}
\end{figure}

We validate our calculated data with the raw IMU data. \BCref{fig:imu_overlay} shows an example of the calculated, target, and measured IMU values in x-axis direction for a sinusoidal motion of frequency $1$ Hz and amplitude $\pm20$ mm/$\pm10$ deg. %

The data preview module showing desired plots and graphs is loaded automatically after the data is processed. This can also be executed independently for quick visualization of the latest data.

\section{Experimental Benchmarking and Results}
In this section, we present benchmarking experiments and results with our 6-DoF motion platform hardware. In order to benchmark the results, we compare the commanded and target values with the calculated and actual values.

We benchmark our setup with two motion trajectories:
\begin{enumerate}
    \item Sinusoidal Trajectory
    \item Circular Trajectory
\end{enumerate}

\subsection{Sinusoidal Trajectory}
We use sinusoidal trajectory as one of the benchmarks because we can test and validate a wide range of dynamic motions with varying amplitude and frequencies on each DoF. We use a target sine trajectory in each of the 6 DoF with a sequence run-time of $3$ seconds, wait-time of $2$ seconds, frequency of $2$ Hz, translational amplitude of $20$ mm and rotational amplitude of $10$ degrees.

\subsubsection{Joint Angles}
\label{sec:sine_joint_angles}
\begin{figure*}[htbp]
    \centerline{\includegraphics{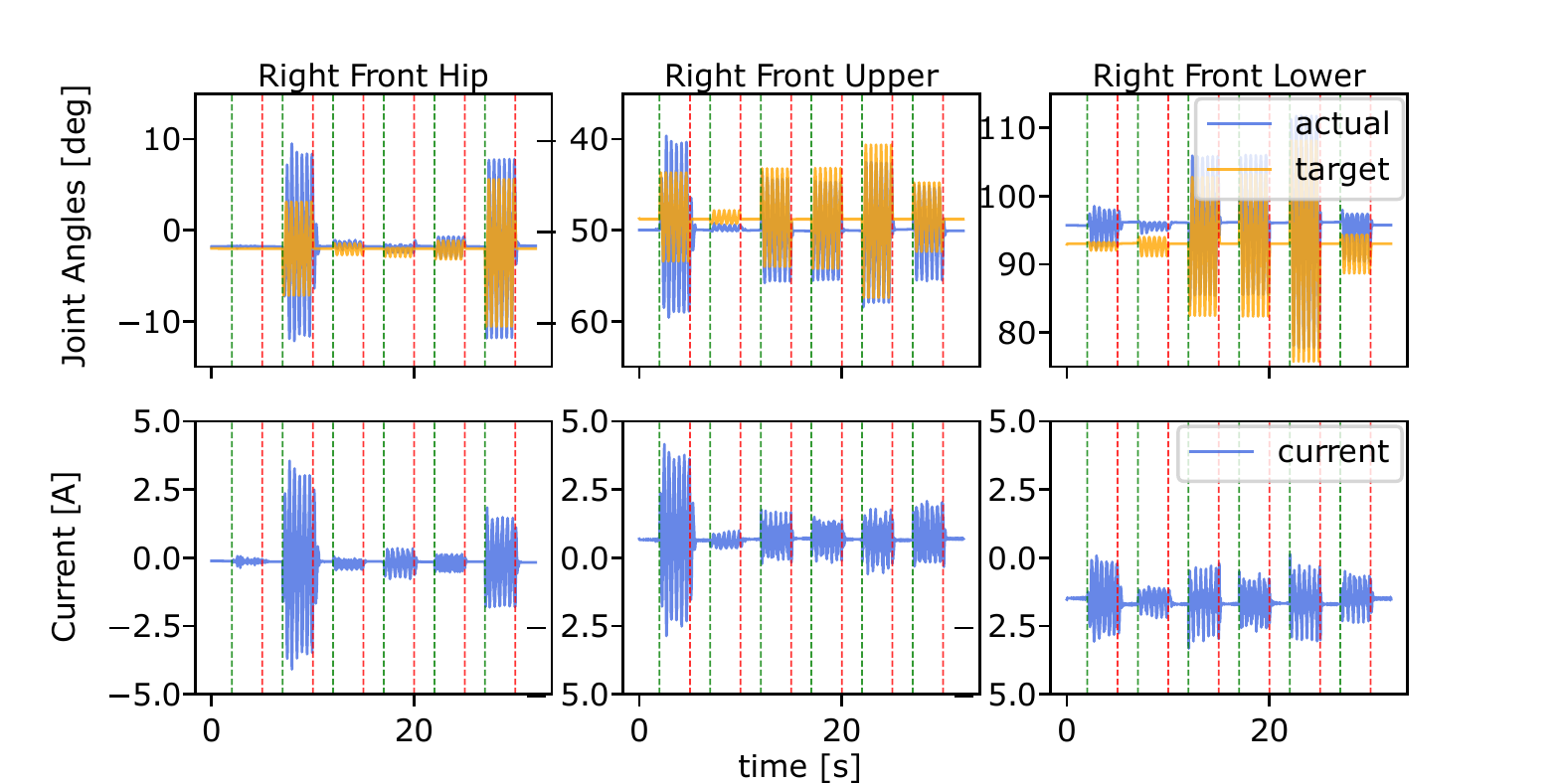}}
    \caption{Robot front right leg target vs actual joint angles for the Solo Control Environment. The respective current measurements are performed in the motor drivers, which is in linear relationship to torque output, with a software limit up to \SI{15}{A}.}
    \label{fig:solo_fr_jointangles}
\end{figure*}

We compare the actual robot joint angles to the target joint angles from the right front leg, and show the current measurement (in linear relation to torque output) of the respective motor \BCref{fig:solo_fr_jointangles}.  %
The Root Mean Squared Error (RMSE) between the target and actual joint angles is calculated as shown in \BCref{tab:msd_jointangles_sine}. 
The average RMSE for the right leg was $2.3$ degrees, showing good tracking results. %
We notice a constant offset in the right front upper and right front lower joint angles, which is caused by gravity. In this control, the influence of gravity is not modeled, and hence not compensated for.

\begin{table}[]
    \centering
    \begin{tabular}{|l|c|}
         \hline\textbf{Joint Name} & \textbf{RMSE} \\ \hline
                                Right Front Hip Joint & $1.6$ deg \\
                                Right Front Upper Joint & $1.9$ deg \\ 
                                Right Front Lower Joint & $3.1$ deg \\\hline
                                Average of Right Leg Joints & $2.3$ deg \\\hline
    \end{tabular}
    \caption{RMSE between target and actual joint angle}
    \label{tab:msd_jointangles_sine}
\end{table}

\subsubsection{Platform Pose}
\begin{figure*}[htbp]
    \centerline{\includegraphics{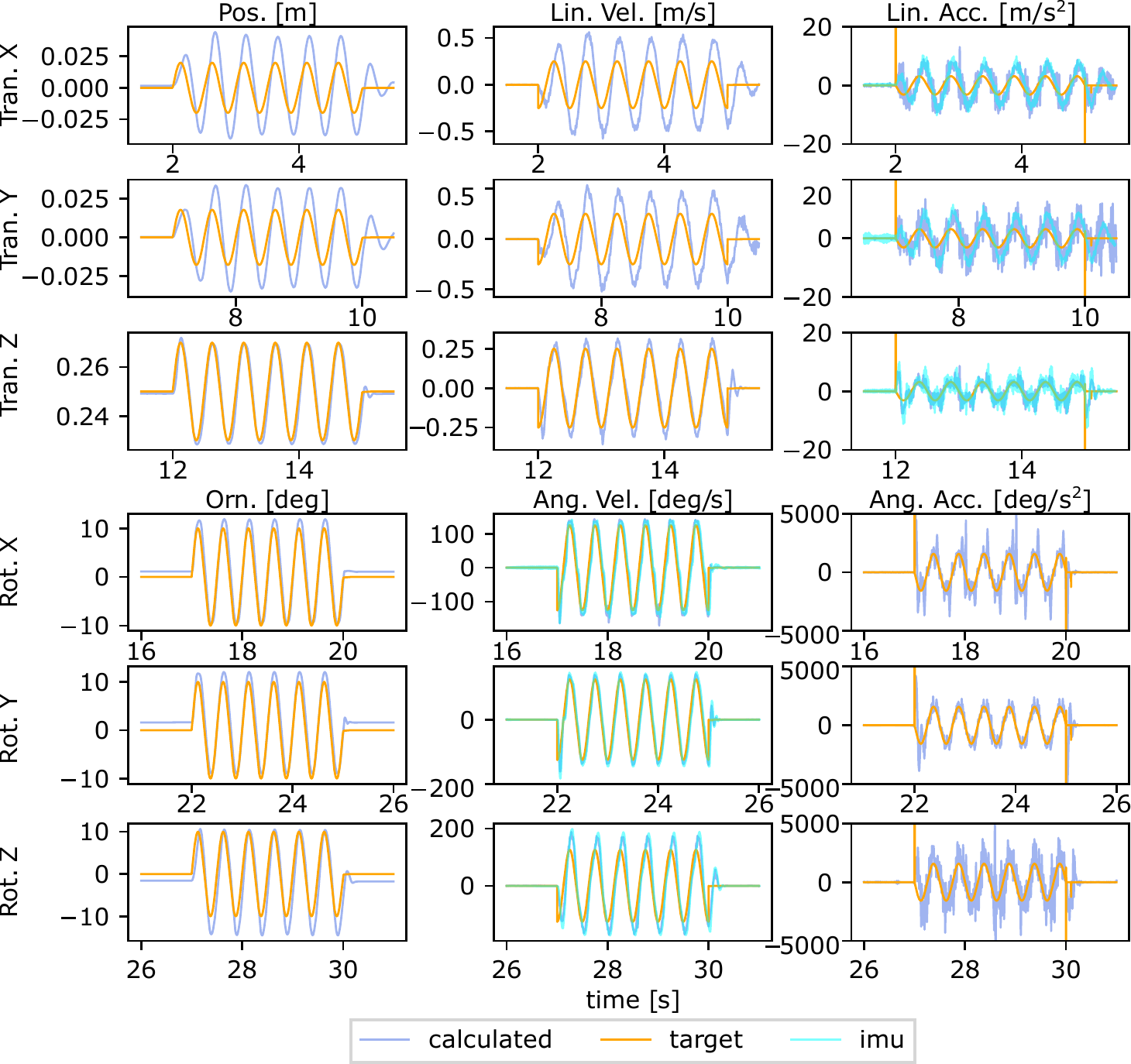}}
    \caption{Motion platform target vs calculated position, linear velocity, linear acceleration, orientation, angular velocity, and angular acceleration for sine trajectory}
    \label{fig:solo_platform_pose}
\end{figure*}

A second approach to benchmark our motion platform is to compare the target pose to the calculated pose (\BCref{sec:post_processing}), as shown in \BCref{fig:solo_platform_pose}.





We observed overshoot motions in x, y translation and in x rotation.
The overshoot motions are associated with the joint overshoots as in \BCref{fig:solo_fr_jointangles}.
The pose tracking RMSE is documented in \BCref{tab:msd_platform_pose_sine}. 
We find that the largest RMSE of $5.8$ mm happens during x-axis translation, followed by $4.7$ mm in the y-axis translation.
We believe fine-tuning the controller gains can improve the tracking performance.

In a set of additional experiments, we increased the platform's sine wave motion frequency to identify the setup's maximum acceleration.
We stopped experiments at \SI{10}{Hz} with an amplitude of \SI{10}{mm}. 
At these loads, the motor current is reaching the limit of \SI{15}{A}.
At \SI{10}{Hz}, the platform movement becomes relatively violent at maximum acceleration of $\approx$\SI{4}{g}, and pose tracking quality decays much.
Nevertheless, this experiment demonstrates the high acceleration capacity of the proposed motion platform.

\begin{table}[]
    \centering
    \begin{tabular}{|l|l|c|c|}
        \hline\textbf{Platform Traj.} & \textbf{Motion Type}  & \textbf{Motion Axis} & \textbf{RMSE} \\ \hline
                              &  & x & $5.8$ mm\\
                              & Translation  & y & $4.7$ mm\\ 
                              &  & z & $1.6$ mm\\\cline{2-4}
        Sine                  & Average Translation & xyz & $4.4$ mm \\\cline{2-4}
        Trajectory            &  & x & $1.3$ deg \\
                              & Rotation & y & $1.6$ deg \\
                              &  & z & $1.9$ deg \\ \cline{2-4}
                              & Average Rotation & xyz & $1.6$ deg \\\hline
    \end{tabular}
    \caption{RMSE Error between Target Platform Pose and Calculated Platform Pose for each axis for Sine Trajectory in Solo Control Environment.}
    \label{tab:msd_platform_pose_sine}
\end{table}

\subsection{Circular Trajectory}
\begin{figure}[htbp]
    \centerline{\includegraphics{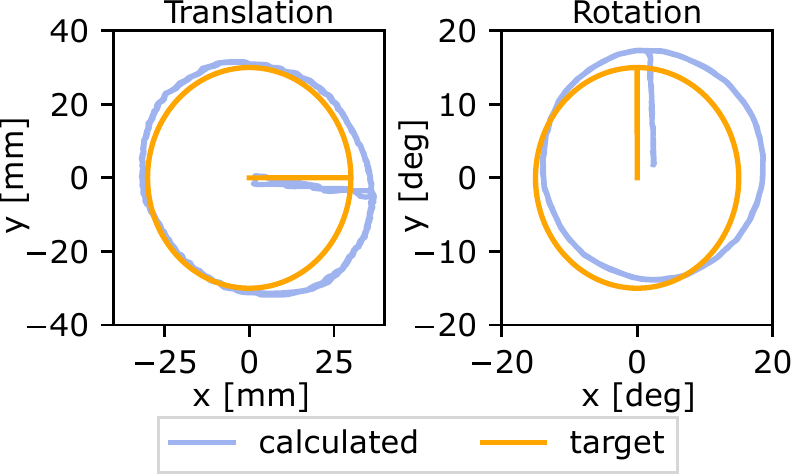}}
    \caption{Motion platform target vs calculated pose for circular trajectory.}
    \label{fig:circular_plot}
\end{figure}

We use a circular trajectory to test dynamic motion tracking of multiple DoF movements simultaneously.
The circular trajectory generation method in \ref{sec:platform_trajectory_generation} to generate a simultaneous clock-wise circular translation and rotation motion on the z-axis in a motion frequency of $2$ Hz (\BCref{fig:circular_plot}). 

The platform can follow the target trajectory as shown in \BCref{fig:circular_plot}.
We notice deviations from the nominal path, which we connect to minor calibration errors.
The platform follows a smooth path consistently over 20 cycles, with a satisfactory and repeatable precision.
The accuracy of path tracking is documented in Table \ref{tab:msd_platform_pose_circ}, which should be improved further.
We also notice relatively high RMSE on z axis translation, which can be improved by implementing a gravity compensation function.

The experiment videos of the motion platform are available at \url{https://youtu.be/thXPA2MYcQw}. 



\begin{table}[]
    \centering
    \begin{tabular}{|l|l|c|c|}
        \hline\textbf{Platform Traj.} & \textbf{Motion Type}  & \textbf{Motion Axis} & \textbf{R.M.S.E} \\ \hline
                              &  & x & $4.0$ mm\\
                              & Translation  & y & $1.6$ mm\\ 
                              &  & z & $7.9$ mm\\\cline{2-4}
        Circular                  & Average Translation & x, y, z & $5.2$ mm \\\cline{2-4}
        Trajectory            &  & x & $2.5$ deg \\
                              & Rotation & y & $1.9$ deg \\
                              &  & z & $1.8$ deg \\ \cline{2-4}
                              & Average Rotation & x, y, z & $2.1$ deg \\\hline
    \end{tabular}
    \caption{RMSE between target and calculated platform pose for each axis for circular trajectory.}
    \label{tab:msd_platform_pose_circ}
\end{table}

\section{Conclusion}
In this work, we propose a high-performance low-cost motion platform design.
The motion platform hardware is centered around the four legs of the quadruped robot Solo. Its legs hold a platform, creating a parallel robot.
Besides the quadruped robot, our setup requires only few, simple-to-implement 3D printing parts, and a few off-the-shell components like bearings and screws.
Driven by the four legs and 12 motors simultaneously, this motion platform provides high acceleration, up to \SI{4}{g} at a platform loading of 300 grams. %
Our systematic benchmarking on sinusoidal and circular trajectories with 1.2 kg platform loading demonstrates good tracking performance and workspace.
We custom-developed a control framework and a PyBullet-based simulation model of the motion platform, allowing for real-time control and data logging up to \SI{1}{kHz}.  %
The hardware design and software packages are fully open source available at \url{https://github.com/nayan-pradhan/solo-6dof-motion-platform}. %
We envision our robot design will support barrier free research, catalyzing future robot development and application.

\section*{Acknowledgments}
The authors thank the International Max Planck Research School for Intelligent Systems (IMPRS-IS) and the China Scholarship Council (CSC) for supporting An Mo.
This work was funded by the Max-Planck Institute for Intelligent Systems' Grassroots project and the Deutsche Forschungsgemeinschaft (DFG, German Research Foundation), project 449912641.
The authors thank Felix Grimminger and Huanbo Sun for supporting the hardware development.

\printbibliography

\end{document}